\documentclass[conference,a4paper]{IEEEtran}
\IEEEoverridecommandlockouts

\usepackage{cite}
\usepackage{amsmath,amssymb,amsfonts}
\usepackage{graphicx}
\usepackage[percent]{overpic}
\usepackage{textcomp}
\usepackage{xcolor}
\usepackage[colorlinks=true, urlcolor=blue, citecolor=blue, linkcolor=darkgray]{hyperref}
\usepackage{cleveref}
\usepackage{booktabs}
\usepackage{eso-pic}

\graphicspath{{figures/}}

\crefname{figure}{Fig.}{Figs.}
\Crefname{figure}{Fig.}{Figs.}
\crefname{table}{Table}{Tables}
\Crefname{table}{Table}{Tables}
\crefname{equation}{Eq.}{Eqs.}
\Crefname{equation}{Eq.}{Eqs.}

\newcommand{\lollypop}{\emph{Lollypop}}
\newcommand{\vect}[1]{\mathbf{#1}}
\newcommand{\norm}[1]{\left\|#1\right\|}

\begin{document}

\title{Lollypop: Camera-to-Motion-Capture Calibration Verification with a Reference Target}

\author{
\IEEEauthorblockN{Tianyi Liu, Kevin Harris, Mihika Dave, Kun He}
\IEEEauthorblockA{\textit{Meta}\\
Redmond, WA, USA\\
Email: \{tianyil, harriskevin, mihikadave, kunhe\}@meta.com}
}

\AddToShipoutPictureFG*{\AtPageLowerLeft{\raisebox{8mm}{\makebox[\paperwidth]{\parbox{0.95\textwidth}{\fontsize{6}{7}\selectfont \copyright\ 2026 IEEE. Personal use of this material is permitted. Permission from IEEE must be obtained for all other uses, in any current or future media, including reprinting/republishing this material for advertising or promotional purposes, creating new collective works, for resale or redistribution to servers or lists, or reuse of any copyrighted component of this work in other works.}}}}}

\maketitle

\begin{abstract}
Camera-to-motion-capture (mocap) calibration is essential for using mocap as ground truth in robotics, AR/VR, and other computer vision tasks.
However, the calibration can drift after deployment, while calibration residuals and visual inspection provide limited independent verification.
We present \lollypop, a fiducial-mocap reference target for independent calibration verification.
The target couples an ArUco fiducial with a mocap marker constellation so the visual center and tracked centroid represent the same physical point.
Given a candidate calibration, verification projects the mocap point into the image and measures its disagreement with the detected fiducial center.
Experiments show sub-pixel nominal error, sensitivity to controlled extrinsic perturbations, and increasing error during an illustrative mixed-handling sequence.
\end{abstract}


\section{Introduction}

Optical motion-capture (mocap) systems are widely used as external reference sensors for robotics, visual-inertial odometry, and virtual reality~\cite{sturm2012tumrgbd,burri2016euroc,schubert2018tumvi}.
Motion-capture data are often used to train and evaluate algorithms that use cameras as the primary sensing modality.
To use data from both modalities, camera observations and mocap measurements must be expressed in a shared coordinate system through calibration.
Camera intrinsic calibration is a well-studied part of the pipeline~\cite{zhang2000flexible,kannala2006generic}.
Camera-to-mocap extrinsic calibration maps between the camera and motion-capture coordinate systems.
Although less frequently discussed, this step has also been addressed by several methods.
A recurring limitation is that prior methods depend on assumptions that can be difficult to maintain or include noise-sensitive substeps.

Calibrating and validating this camera-to-mocap chain is difficult because the two systems do not naturally observe the same feature.
Optical mocap systems commonly localize retroreflective markers illuminated by infrared light using a calibrated array of mocap cameras.
Device cameras, by contrast, observe the scene through their own imaging pipeline and are typically used to detect fiducials, texture, or object features rather than mocap marker centroids.
Thus, a 3D point reported by mocap does not automatically correspond to a precise image point, and vice versa.

\begin{figure}[t]
  \centering
  \includegraphics[page=1,width=\linewidth]{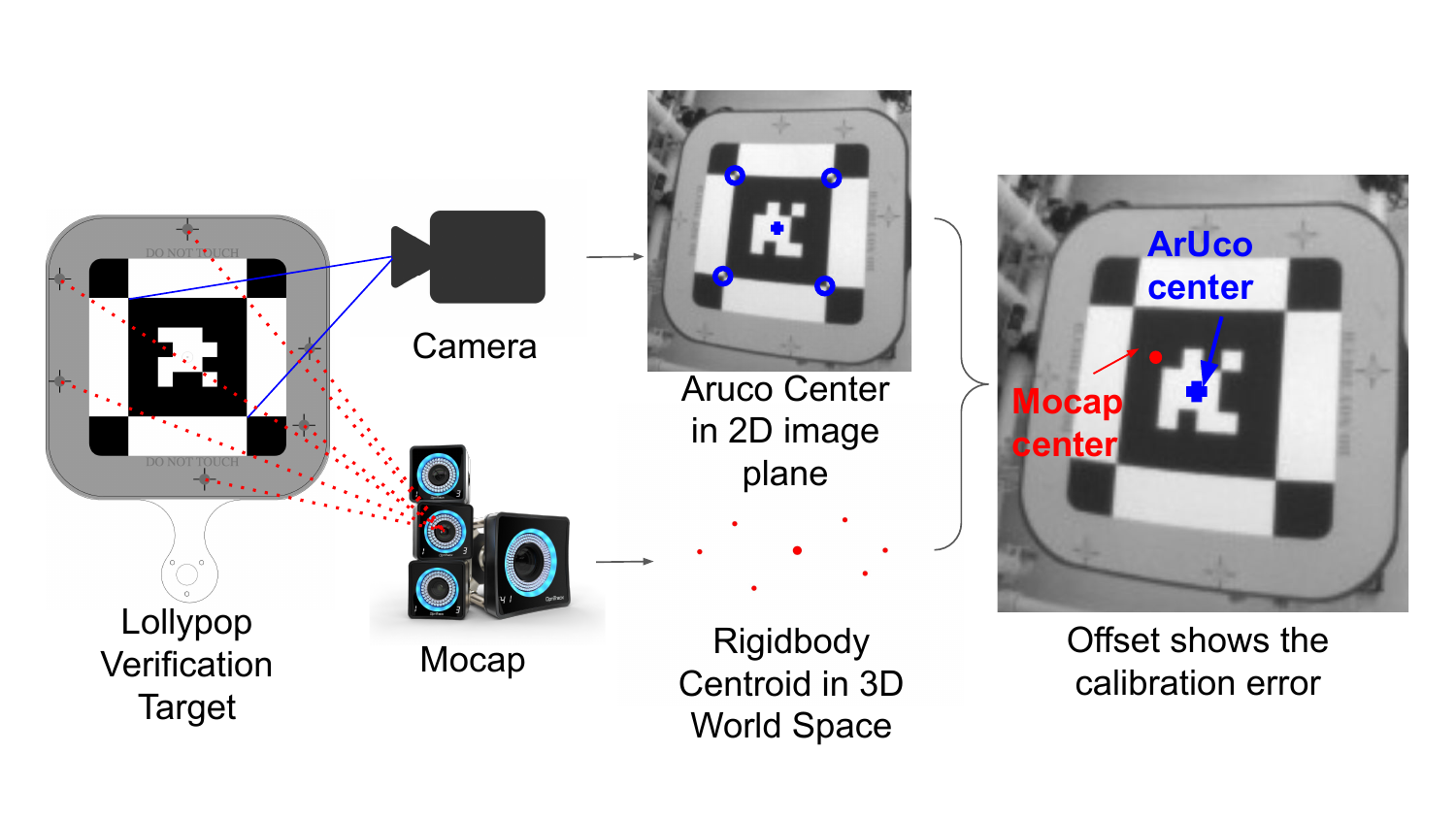}
  \caption{System overview of \lollypop{} calibration verification. The target creates a shared physical reference point between the camera image and the motion-capture system; verification compares the detected fiducial center with the mocap centroid projected through the candidate camera-to-mocap calibration.}
  \label{fig:system_overview}
\end{figure}

Prior work has tried to address camera-to-mocap calibration from several directions.
Schofield et al.~\cite{schofield2018calibration} use bundle adjustment to estimate camera-motion-capture extrinsics and evaluate them with 3D marker-position error and 2D reprojection error.
Their 3D metric can be biased by manual marker placement on the checkerboard because the assumed marker offsets may not match the true physical marker centers.
Their 2D metric depends on detecting retroreflective markers as bright blobs using thresholding and circle fitting, which can introduce error under changes in illumination, saturation, viewing angle, occlusion, and threshold choice.
Chiodini et al.~\cite{chiodini2018camera} calibrate a camera rig using projected mocap markers and include uncertainty analysis, but the image-side observation is still a thresholded bright-blob centroid that can suffer from the same issues as~\cite{schofield2018calibration}.
Ganesh et al.~\cite{ganesh2021extrinsic} evaluate transform-chain consistency by comparing a predicted fiducial-board pose against a mocap-measured board pose, yielding a transform-space consistency error rather than a direct image-space test of the final projection.
Datasets and benchmarks such as TUM RGB-D, EuRoC, and TUM VI demonstrate the importance of calibrated ground truth~\cite{sturm2012tumrgbd,burri2016euroc,schubert2018tumvi}.
However, their publications provide limited detail on independent verification of the camera-to-ground-truth alignment.

In this paper, we focus on post-calibration verification through two contributions, summarized in \cref{fig:system_overview}.
First, we present a hardware reference target, the \lollypop{} target, that mechanically co-registers a camera-visible fiducial center with a mocap-tracked centroid, reducing the risk that handling changes their physical offset.
Second, we present an image-space verification procedure that avoids intensity-thresholded retroreflective-marker localization in the device image.
We evaluate the target under a nominal calibration, controlled extrinsic perturbations, and an illustrative mixed-handling sequence.

\section{Target Design}

\begin{figure}[t]
  \centering
  \includegraphics[width=\linewidth]{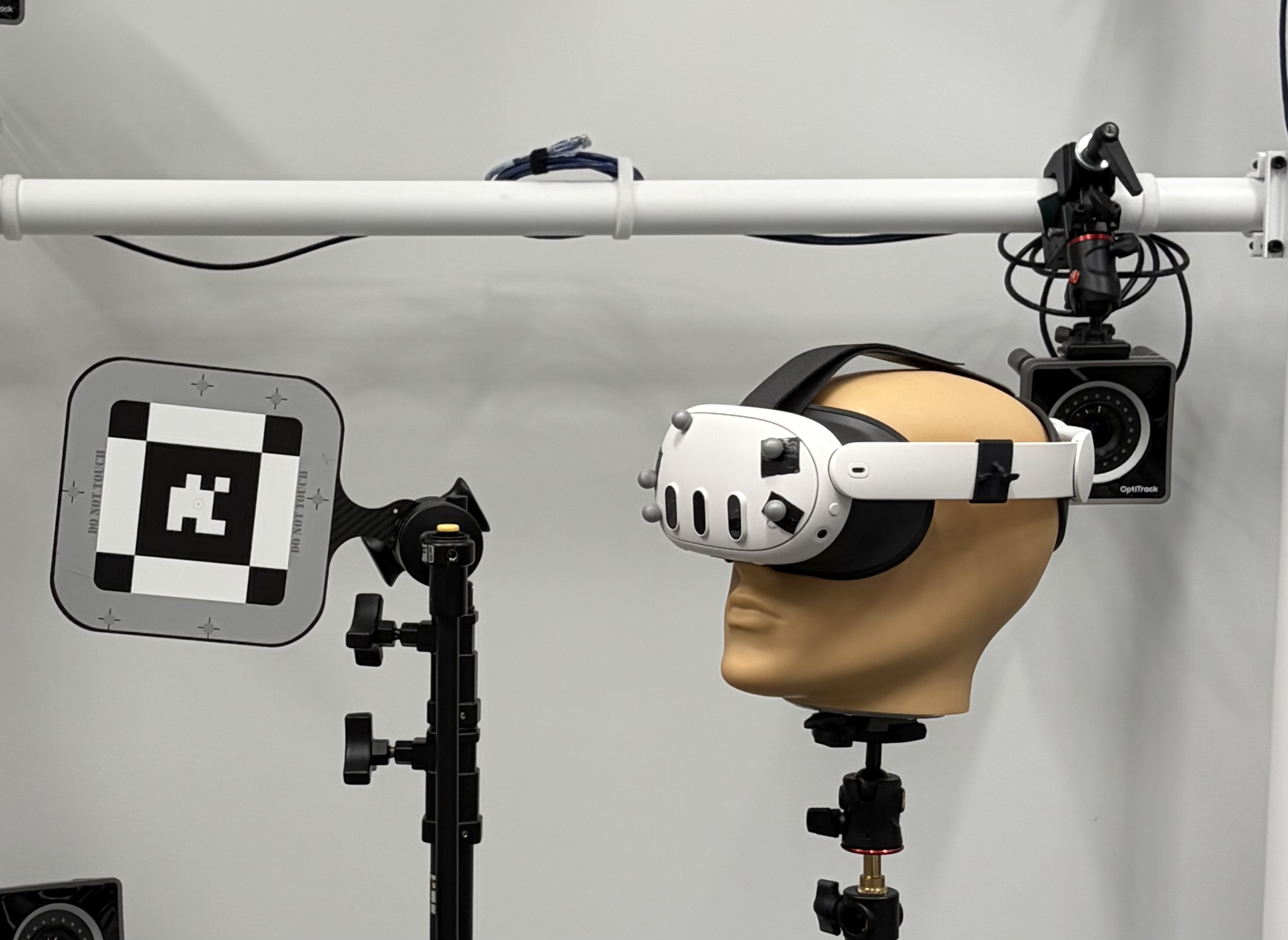}
  \caption{The \lollypop{} reference target. An ArUco fiducial and six retroreflective markers, with the marker centroid aligned to the fiducial center.}
  \label{fig:target}
\end{figure}

As shown in Fig.~\ref{fig:target}, the target combines an ArUco fiducial marker~\cite{garrido2014automatic,romero2018speeded} with six retroreflective marker stickers mounted on the same rigid planar backing.
The ArUco marker provides an image-space center estimate from detected marker corners.
The six mocap markers are placed asymmetrically to avoid orientation ambiguity in mocap tracking.
The six-marker constellation is redundant relative to the three markers required to define a rigid body, improving tracking robustness in OptiTrack Motive under occlusion and marker dropout.
The printed target is attached to a carbon-fiber plate to maximize rigidity and keep the fiducial plane and marker constellation fixed during handling.
Critically, the device is constructed so that the ArUco board center and the mocap rigid body centroid are coincident by construction.
This is a fixed mechanical constraint of the rigid device. Unlike prior methods that require human placement or measurement, this design removes the need for manual alignment.

This planar design simplifies manufacturing relative to a multi-planar marker arrangement. The primary source of manufacturing error is the thickness of the retroreflective tape (0.12\,mm), which raises the mocap marker centroids above the printed board plane.
In practice, ArUco detection is inherently limited to moderate viewing angles, bounding this error well below 0.12\,mm; additional filtering on the detection viewing angle can further reduce its impact for tighter tolerance requirements.
Because the mocap system triangulates each marker using multiple cameras, coplanarity does not make scale or depth unobservable.

\section{Methods}

\subsection{Problem Definition and Notation}
A 3D point $\vect{X}$ in the world frame is projected into a camera $c$ mounted on a mocap-tracked rigid body via the following chain:
\begin{equation}
    \vect{u}_{c} = \pi\!\left(
    T^{\mathrm{c}}_{\mathrm{rb}}\,
    T^{\mathrm{rb}}_{\mathrm{w}}\,
    \vect{X}
    ;\; \boldsymbol{\theta}_c\right)
    \label{eq:kinematic_chain}
\end{equation}
where:
\begin{itemize}\setlength{\itemsep}{3pt}
    \item $T^{\mathrm{rb}}_{\mathrm{w}}$ is the mocap-reported rigid body pose,
    \item $T^{\mathrm{c}}_{\mathrm{rb}}$ is the camera extrinsic, mapping rigid body coordinates into the camera frame,
    \item $\pi(\cdot; \boldsymbol{\theta}_c)$ is the camera projection function that produces pixel coordinates using intrinsic parameters $\boldsymbol{\theta}_c$.
\end{itemize}

\subsection{Error Metrics}

For each frame where both the ArUco marker and the mocap rigid body are detected on the target, we compare two cross-modal estimates of the device center in the image (\cref{fig:system_overview}) in both 2D and 3D.

\paragraph{ArUco center}
The board center is estimated from the four detected corners via a 2D homography. Each corner is undistorted to normalized image coordinates using the camera model.
A homography $\vect{H}$ from a canonical unit square to these normalized positions is estimated,
and the canonical center $(0,0)$ is warped through $\vect{H}$ and mapped back to distorted pixel coordinates to yield $\vect{u}_\text{aruco}$. This operates entirely in the image domain without 3D pose estimation.
A full ArUco-pose comparison is useful for 6-DoF diagnosis; the center-only metric directly tests image registration without printed-scale or planar pose-estimation errors, but does not independently diagnose depth or rotation.

\paragraph{Mocap center}
We obtain the 3D \lollypop{} rigid-body centroid $\vect{X}_\text{mocap}$ directly from the mocap tracking system in the world frame, and project it through the chain of \cref{eq:kinematic_chain} to predict its image location:
\begin{equation}
    \vect{u}_\text{mocap} = \pi\!\left(T^{\mathrm{c}}_{\mathrm{rb}}\,T^{\mathrm{rb}}_{\mathrm{w}}\,\vect{X}_\text{mocap};\; \boldsymbol{\theta}_c\right).
    \label{eq:mocap_proj}
\end{equation}

\paragraph{2D error}
The primary metric is the pixel-space distance between the two image estimates:
\begin{equation}
  e_{2D} = \norm{\vect{u}_\text{aruco} - \vect{u}_\text{mocap}}_2.
  \label{eq:e2d}
\end{equation}
Thus, $e_{2D}$ measures consistency of the complete projection chain and may reflect errors in the candidate extrinsic, mocap poses, camera intrinsics, synchronization, target construction, and ArUco localization; it does not isolate extrinsic error alone.

\paragraph{3D error}
We also report a metric-space error. Let $\vect{X}^{\mathrm{c}}_\text{mocap} = T^{\mathrm{c}}_{\mathrm{rb}}\,T^{\mathrm{rb}}_{\mathrm{w}}\,\vect{X}_\text{mocap}$ be the mocap centroid in the camera frame, and let $\vect{X}^{\mathrm{c}}_\text{aruco} = \pi^{-1}\!\left(\vect{u}_\text{aruco},\, d;\, \boldsymbol{\theta}_c\right)$ be the ArUco center back-projected to depth $d = \big[\vect{X}^{\mathrm{c}}_\text{mocap}\big]_z$. The error is
\begin{equation}
  e_{3D} = \norm{\vect{X}^{\mathrm{c}}_\text{aruco} - \vect{X}^{\mathrm{c}}_\text{mocap}}_2.
  \label{eq:e3d}
\end{equation}
Note that this value is not fully independent in the depth direction, since the back-projection depth is derived from mocap.
This metric is useful for geometric intuition, and we treat $e_{2D}$ as the primary metric.

\section{Experiments}

\subsection{Experimental Setup}

We evaluate the target using a Meta Quest 3 headset with four fisheye cameras inside a 36-camera OptiTrack Prime 41 motion-capture volume.
Five mocap markers mounted on the headset define the tracked headset rigid body, while the \lollypop{} target is tracked as a separate rigid body.
Camera intrinsics are held fixed from device calibration, and candidate camera-to-mocap extrinsics are obtained from a prior calibration step.
Headset camera frames are hardware synchronized to mocap measurements at exposure time.
A full \lollypop{} verification recording consists of a human operator waving the \lollypop{} target in front of each camera at the desired distance and covering the whole image plane.

\subsection{Calibration Quality Verification}

For the first experiment, we evaluated a nominal calibration expected to produce low projection error.

To generate the final results, we collected five recordings, each containing about 800 valid frames.
We make no distributional assumption and therefore emphasize the empirical median and 95th percentile alongside RMSE.
\Cref{tab:percamera} reports the per-camera \lollypop{} verification error for the nominal calibration on our four-camera headset.
The combined median and RMSE are sub-pixel ($0.53$~px and $0.80$~px), with a 95th percentile of $1.47$~px, corresponding to a combined metric error of $0.84$~mm median ($2.02$~mm at p95).

To evaluate sensitivity to controlled extrinsic perturbations, we re-scored the same recorded frames while perturbing the candidate extrinsic $T^{\mathrm{c}}_{\mathrm{rb}}$ by a known offset.
We injected pure translation offsets of $1$ and $5$~mm and pure rotation offsets of $0.2$ and $1.0$\textdegree{} to each axis.
\Cref{tab:sensitivity} reports the resulting combined $e_{2D}$.
The verification error increases with perturbation magnitude.
The smallest injected errors ($1$~mm and $0.2$\textdegree{}) already raise the RMSE to $1.43$ and $1.50$~px---near the baseline 95th percentile of $1.47$~px and roughly double the baseline RMSE of $0.80$~px---while $5$~mm and $1.0$\textdegree{} errors raise it to $5.63$ and $6.46$~px, about $7$--$8\times$ the baseline.
Thus, \lollypop{} responds measurably to the tested $1$~mm and $0.2$\textdegree{} perturbations, while the larger perturbations produce correspondingly larger errors.

\begin{table}[t]
  \centering
  \caption{Per-camera \lollypop{} pixel error $e_{2D}$ for the nominal calibration.}
  \label{tab:percamera}
  \footnotesize
  \setlength{\tabcolsep}{4pt}
  \begin{tabular}{@{}lrrrrrr@{}}
    \toprule
    \multicolumn{7}{c}{Pixel error $e_{2D}$ (px)}\\
    \midrule
    Camera & RMSE & Mean & Med. & Std & p95 & Max\\
    \midrule
    Front Left  & 0.48 & 0.36 & 0.25 & 0.32 & 1.12 & 1.33\\
    Front Right & 0.62 & 0.47 & 0.36 & 0.41 & 1.55 & 1.86\\
    Side Left   & 0.97 & 0.88 & 0.80 & 0.42 & 1.48 & 1.62\\
    Side Right  & 1.01 & 0.93 & 0.99 & 0.39 & 1.50 & 1.74\\
    \midrule
    Combined    & 0.80 & 0.65 & 0.53 & 0.46 & 1.47 & 1.86\\
    \bottomrule
  \end{tabular}
\end{table}

\begin{table}[t]
  \centering
  \caption{\lollypop{} pixel error $e_{2D}$ (px) under controlled errors injected into the candidate extrinsic $T^{\mathrm{c}}_{\mathrm{rb}}$.}
  \label{tab:sensitivity}
  \footnotesize
  \setlength{\tabcolsep}{4pt}
  \begin{tabular}{@{}lrrrrrr@{}}
    \toprule
    Perturbation & RMSE & Mean & Med. & Std & p95 & Max\\
    \midrule
    Baseline (none)                   & 0.80 & 0.65 & 0.53 & 0.46 & 1.47 & 1.86\\
    \midrule
    $\Delta t = 1$~mm                 & 1.43 & 1.29 & 1.19 & 0.62 & 2.43 & 3.41\\
    $\Delta t = 5$~mm                 & 5.63 & 5.28 & 5.53 & 1.95 & 8.06 & 10.48\\
    \midrule
    $\Delta\theta = 0.2$\textdegree{} & 1.50 & 1.44 & 1.45 & 0.40 & 2.11 & 2.63\\
    $\Delta\theta = 1.0$\textdegree{} & 6.46 & 6.12 & 6.63 & 2.06 & 8.47 & 8.76\\
    \bottomrule
  \end{tabular}
\end{table}

\subsection{Detecting Degradation}

\begin{figure}[t]
  \centering
  \includegraphics[width=0.88\linewidth]{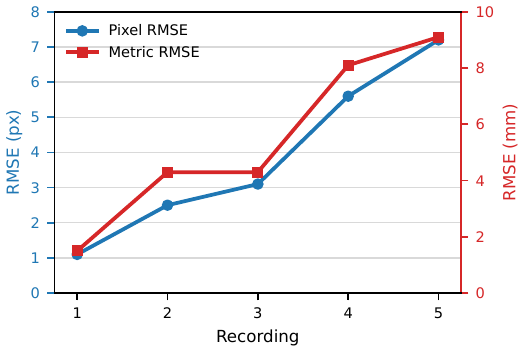}
  \caption{Verification error during an illustrative mixed-handling sequence involving headset repositioning and occasional marker contact.}
  \label{fig:drift}
\end{figure}

Calibration quality can degrade after deployment as devices are handled, mounted, or perturbed.
It is often difficult to know when the system needs to be re-calibrated, and practitioners often rely on visual inspection of the marker reprojection.
To test whether the target exposes this degradation,
we collected verification recordings after an initial calibration while repeatedly donning and doffing the headset and occasionally bumping the headset mocap markers to simulate real-world handling.
Incidental marker displacement or mount flex can change the markers' relationship to the cameras, invalidating the original calibration and increasing verification error.
\Cref{fig:drift} shows that both image-space and metric-space errors increase over the sequence, indicating that the target can act as a session-level calibration health check rather than a one-time calibration residual.

\section{Conclusion}

We presented \lollypop, a fiducial-mocap reference target that co-registers a camera-visible fiducial center with a mocap-tracked centroid, together with a post-calibration image-space verification procedure.
The procedure avoids threshold-based bright-blob localization in the device image and scores consistency of the complete projection chain rather than reporting isolated extrinsic parameters.
Experiments on a multi-camera fisheye headset show sub-pixel nominal error, sensitivity to controlled extrinsic perturbations, and increasing complete-chain inconsistency during an illustrative mixed-handling sequence.
Future work includes multi-point or non-coplanar reference targets and stronger depth observability.

\end{document}